\documentclass[11pt,a4paper]{article}
\usepackage{authblk}
\usepackage[hyperref]{acl2019}

\usepackage{amsmath}
\usepackage{times}
\usepackage{latexsym}
\usepackage{verbatim}
\usepackage{booktabs}
\usepackage{url}

\newcommand{\Sref}[1]{\S\ref{#1}}

\newcommand{\Tref}[1]{Table~\ref{#1}}

\newcommand{\ignore}[1]{}

\aclfinalcopy 
\def\aclpaperid{006} 

\title{Measuring Bias in Contextualized Word Representations}

\author{Keita Kurita ~ Nidhi Vyas ~ Ayush Pareek ~ Alan W Black ~ Yulia Tsvetkov\\
Carnegie Mellon University \\
{\tt \{kkurita,nkvyas,apareek,awb,ytsvetko\}@andrew.cmu.edu}}

\date{}

\begin{document}
\maketitle

\begin{abstract} 
Contextual word embeddings such as BERT have achieved state of the art performance in numerous NLP tasks. Since they are optimized to capture the statistical properties of training data, they tend to pick up on and amplify social stereotypes present in the data as well. In this study, we (1)~propose a template-based method to quantify bias in BERT; (2)~show that this method obtains more consistent results in capturing social biases than the traditional cosine based method; and (3)~conduct a case study, evaluating gender bias in a downstream task of Gender Pronoun Resolution. Although our case study focuses on gender bias, the proposed technique is generalizable to unveiling other biases, including in multiclass settings, such as racial and religious biases.


\end{abstract}

\section{Introduction} 
Type-level word embedding models, including word2vec and GloVe \cite{mikolov2013distributed,pennington2014glove}, have been shown to exhibit social biases present in human-generated training data \cite{bolukbasi2016man,caliskan2017semantics,garg2018word,manzini19multiclass}.  These embeddings are then used in a plethora of downstream applications, which perpetuate and further amplify stereotypes \cite{zhao2017men,leino2018feature}. 
To reveal and quantify corpus-level biases is word embeddings, \citet{bolukbasi2016man} used the word analogy task \cite{mikolov2013distributed}. For example, they showed that gendered male word embeddings like \textit{he, man} are associated with higher-status jobs like \textit{computer programmer} and \textit{doctor}, whereas gendered 
words like \textit{she} or \textit{woman} are associated with \textit{homemaker} and \textit{nurse}. 

Contextual word embedding models, such as ELMo and BERT \cite{peters2018deep,devlin2018bert} have become increasingly common, replacing traditional type-level embeddings and attaining new state of the art results in the majority of NLP tasks. In these models, every word has a different embedding, depending on the context and the language model state; in these settings, the analogy task used to reveal biases in uncontextualized embeddings is not applicable. 
Recently, \citet{may2019measuring} showed that traditional cosine-based methods for exposing bias in sentence embeddings fail to produce consistent results for embeddings generated using contextual methods. 
We find similar inconsistent results with cosine-based methods of exposing bias; this is a motivation to the development of a novel bias test that we propose.

In this work, we propose a new method to quantify bias in BERT embeddings (\Sref{sec:method}). 
Since BERT embeddings use a \emph{masked} language modelling objective, we directly query the model to measure the bias for a particular token. More specifically, we create simple template sentences containing the attribute word for which we want to measure bias (e.g. \emph{programmer}) and the target for bias (e.g. \emph{she} for gender). We then mask the attribute and target tokens sequentially, to get a relative measure of bias across target classes (e.g. male and female). Contextualized word embeddings for a given token change based on its context, so such an approach allows us measure the bias for similar categories divergent by the the target attribute (\Sref{sec:method}). 
We compare our approach with the cosine similarity-based approach (\Sref{sec:corr_human_biases}) and show that our measure of bias is more consistent with human biases and is sensitive to a wide range of biases in the model using various stimuli presented in \citet{caliskan2017semantics}. 
Next, we investigate the effect of a specific type of bias in a specific downstream task: gender bias in BERT and its effect on the task of Gendered Pronoun Resolution (GPR) \citep{webster2018gap}. We show that the bias in GPR is highly correlated with our measure of bias (\Sref{sec:case_study}). Finally, we highlight the potential negative impacts of using BERT in downstream real world applications (\Sref{sec:implications}). The code and data used in this work are publicly available.\footnote{\url{https://bit.ly/2EkJwh1}}



\section{Quantifying Bias in BERT} 
\label{sec:method}
BERT is trained using a masked language modelling objective i.e. to predict masked tokens, denoted as [MASK], in a sentence given the entire context. We use the predictions for these [MASK] tokens to measure the bias encoded in the actual representations. 

We directly query the underlying masked language model in BERT\footnote{For all experiments we use the uncased version of BERT\textsubscript{BASE} \url{https://storage.googleapis.com/bert_models/2018_10_18/uncased_L-12_H-768_A-12.zip}. } to compute the association between certain \textbf{targets} (e.g., gendered words) and \textbf{attributes} (e.g. career-related words). For example, to compute the association between the target \textit{male gender} and the attribute \textit{programmer}, we feed in the masked sentence ``[MASK] is a programmer'' to BERT, and compute the probability assigned to the sentence `\textit{he} is a programmer'' ($p_{tgt}$). To measure the association, however, we need to measure how much \emph{more} BERT prefers the male gender association with the attribute \textit{programmer}, compared to the female gender. We thus re-weight this likelihood $p_{tgt}$ using the prior bias of the model towards predicting the male gender. To do this, we mask out the attribute \textit{programmer} and query BERT with the sentence ``[MASK] is a [MASK]'', then compute the probability BERT assigns to the sentence `\textit{he} is a [MASK]'' ($p_{prior}$). Intuitively, $ p_{prior} $ represents how likely the word \textit{he} is in BERT, given the sentence structure and no other evidence. Finally, the difference between the normalized predictions for the words \textit{he} and \textit{she} can be used to measure the gender bias in BERT for the \textit{programmer} attribute.

Generalizing, we use the following procedure to compute the association between a target and an attribute:
\begin{enumerate}
\itemsep-0.5em 
    \item Prepare a template sentence \\ e.g.``[TARGET] is a [ATTRIBUTE]"
    \item Replace [TARGET] with [MASK] and compute $p_{tgt}$=P([MASK]=[TARGET]$|$ sentence)
    \item Replace both [TARGET] and [ATTRIBUTE] with [MASK], and compute prior probability $p_{prior}$=P([MASK]=[TARGET]$|$ sentence) 
    \item Compute the association as $\log{\frac{p_{tgt}}{p_{prior}}}$ 
\end{enumerate}

We refer to this normalized measure of association as the \emph{increased log probability} score and the difference between the increased log probability scores for two targets (e.g. he/she) as \emph{log probability bias score} which we use as measure of bias. Although this approach requires one to construct a template sentence, these templates are merely simple sentences containing attribute words of interest, and can be shared across multiple targets and attributes.  Further, the flexibility to use such templates can potentially help measure more fine-grained notions of bias in the model.

In the next section, we show that our proposed \emph{log probability bias score} method is more effective at exposing bias than traditional cosine-based measures. 





\begin{table*}
\small
\begin{center}
\begin{tabular}{lp{18em}}
\hline Category & Templates \\ \hline
Pleasant/Unpleasant (Insects/Flowers) & T are A, T is A \\
Pleasant/Unpleasant (EA/AA) & T are A, T is A \\
Career/Family (Male/Female) & T likes A, T like A, T is interested in A \\
Math/Arts (Male/Female) & T likes A, T like A, T is interested in A \\
Science/Arts (Male/Female) & T likes A, T like A, T is interested in A  \\
\hline
\end{tabular}
\end{center}
\caption{\label{weat-templates1} Template sentences used for the WEAT tests (T: target, A: attribute)}
\end{table*}

\begin{table*}
\small
\begin{center}
\begin{tabular}{llp{18em}}
\hline Category & Targets & Templates \\ \hline
Pleasant/Unpleasant (Insects/Flowers) & flowers,insects,flower,insect & T are A, the T is A \\
Pleasant/Unpleasant (EA/AA) & black, white & T people are A, the T person is A \\
Career/Family (Male/Female) & he,she,boys,girls,men,women & T likes A, T like A, T is interested in A \\
Math/Arts (Male/Female) & he,she,boys,girls,men,women & T likes A, T like A, T is interested in A \\
Science/Arts (Male/Female) & he,she,boys,girls,men,women & T likes A, T like A, T is interested in A  \\
\hline
\end{tabular}
\end{center}
\caption{\label{weat-templates2} Template sentences used and target words for the grammatically correct sentences (T: target, A: attribute)}
\end{table*}

\begin{table*}
\small
\begin{center}
\begin{tabular}{lrrr}
\toprule Category & WEAT on GloVe & WEAT on BERT & Ours on BERT  \\ 
& & & \textit{Log Probability Bias Scor}e \\ \midrule
Pleasant/Unpleasant (Insects/Flowers) & 1.543* & 0.6688 & 0.8744*  \\
Pleasant/Unpleasant (EA/AA) & 1.012  & 1.003 & 0.8864*  \\
Career/Family (Male/Female) & 1.814* & 0.5047 & 1.126* \\
Math/Arts (Male/Female) & 1.061 & 0.6755 & 0.8495* \\
Science/Arts (Male/Female) & 1.246*  & 0.8815 & 0.9572* \\
\bottomrule
\end{tabular}
\end{center}
\caption{\label{tab:weat} Effect sizes of bias measurements on WEAT Stimuli. (* indicates significant at $p<0.01$)}
\end{table*}

\section{Correlation with Human Biases}
\label{sec:corr_human_biases}
We investigate the correlation between our measure of bias and human biases. To do this, we apply the log probability bias score to the same set of attributes that were shown to exhibit human bias in experiments that were performed using the Implicit Association Test \cite{iat}. Specifically, we use the stimuli used in the Word Embedding Association Test (WEAT) \cite{caliskan2017semantics}.

\noindent\textbf{Word Embedding Association Test (WEAT)}: The WEAT method compares set of target concepts (e.g. male and female words) denoted as $X$ and $Y$ (each of equal size $N$), with a set of attributes to measure bias over social attributes and roles (e.g. career/family words) denoted as $A$ and $B$. The degree of bias for each target concept $t$ is calculated as follows:  
\begin{align*}
\small
s(t, A, B) = [\textrm{mean}_{a \in A}{\textrm{sim}(t, a)} -  \nonumber
    \textrm{mean}_{b \in B}{\textrm{sim}(t, b)}], 
\end{align*}
where \textit{sim} is the cosine similarity between the embeddings. The test statistics is
\begin{align*}
\small
S(X, Y, A, B) = [\textrm{mean}_{x \in X}s(x, A, B) - \\ \small
    \textrm{mean}_{y \in Y}s(y, A, B)], 
\end{align*}
where the test is a permutation test over $X$ and $Y$. The $p$-value is computed as
\begin{align*}
\small
p = \Pr[S(X_i, Y_i, A, B) > S(X, Y, A, B)] 
\end{align*}
The effect size is measured as
$$ d = \frac{S(X, Y, A, B)}{\textrm{std}_{t \in X \cup Y} s(t, A, B) } $$
It is important to note that the statistical test is a permutation test, and hence a large effect size does not guarantee a higher degree of statistical significance.

\subsection{Baseline: WEAT for BERT} To apply the WEAT method on BERT, we first compute the embeddings for target and attribute words present in the stimuli using multiple templates, such as ``TARGET is ATTRIBUTE" (Refer \Tref{weat-templates1} for an exhaustive list of templates used for each category). We mask the TARGET to compute the embedding\footnote{We use the outputs from the final layer of BERT as embeddings} for the ATTRIBUTE and vice versa. Words that are absent in the BERT vocabulary are removed from the targets. We ensure that the number of words for both targets are equal, by removing random words from the smaller target set. To confirm whether the reduction in vocabulary results in a change of $p$-value, we also conduct the WEAT on GloVe with the reduced vocabulary.\footnote{WEAT was originally used to study the GloVe embeddings}

\subsection{Proposed: Log Probability Bias Score}
To compare our method measuring bias, and to test for human-like biases in BERT, we also compute the \emph{log probability bias score} for the same set of attributes and targets in the stimuli. We compute the mean \emph{log probability bias score} for each attribute, and permute the attributes to measure statistical significance with the permutation test.  Since many TARGETs in the stimuli cause the template sentence to become grammatically incorrect, resulting in low predicted probabilities, we fixed the TARGET to common pronouns/indicators of category such as \textit{flower, he, she} (\Tref{weat-templates2} contains a full list of target words and templates). This avoids large variance in predicted probabilities, leading to more reliable results. The effect size is computed in the same way as the WEAT except the standard deviation is computed over the mean \emph{log probability bias scores}.

We experiment over the following categories of stimuli in the WEAT experiments: Category 1 (flower/insect targets and pleasant/unpleasant attributes), Category 3 (European American/African American names and pleasant/unpleasant attributes), Category 6 (male/female names and career/family attributes), Category 7 (male/female targets and math/arts attributes) and Category 8 (male/female targets and science/arts attributes).

\subsection{Comparison Results}
The WEAT on GloVe returns similar findings to those of \citet{caliskan2017semantics} except for the European/African American names and pleasant/unpleasant association not exhibiting significant bias. This is due to only 5 of the African American names being present in the BERT vocabulary. The WEAT for BERT fails to find any statistically significant biases at $p<0.01$. This implies that WEAT is not an effective measure for bias in BERT embeddings, or that methods for constructing embeddings require additional investigation. In contrast, our method of querying the underlying language model exposes statistically significant association across all categories, showing that BERT does indeed encode biases and that our method is more sensitive to them.




\section{Case Study: Effects of Gender Bias on Gendered Pronoun Resolution}
\label{sec:case_study}
\paragraph{Dataset}
We examined the downstream effects of bias in BERT using the Gendered Pronoun Resolution (GPR) task \cite{webster2018gap}. GPR is a sub-task in co-reference resolution, where a pronoun-containing expression is to be paired with the referring expression. Since pronoun resolving systems generally favor the male entities \cite{webster2018gap}, this task is a valid test-bed for our study. We use the GAP dataset\footnote{\url{https://github.com/google-research-datasets/gap-coreference}} by \citet{webster2018gap}, containing 8,908 human-labeled ambiguous pronoun-name pairs, created from Wikipedia. The task is to classify whether an ambiguous pronoun $P$ in a text refers to entity $A$, entity $B$ or neither. There are 1,000 male and female pronouns in the training set each, with 103 and 98 of them not referring to any entity in the sentence, respectively. 


\paragraph{Model} We use the model suggested on Kaggle,\footnote{\url{https://www.kaggle.com/mateiionita/taming-the-bert-a-baseline}} inspired by \citet{47786}. The model uses BERT embeddings for $P$, $A$ and $B$, given the context of the input sentence. Next, it uses a multi-layer perceptron (MLP) layer to perform a naive classification to decide if the pronoun belongs to $A$, $B$ or neither. The MLP layer uses a single hidden layer with 31 dimensions, a dropout of 0.6 and L2 regularization with weight 0.1. 

\begin{table}
\small
\centering
\begin{tabular}{lrr}
\toprule
              Gender             & Prior Prob. & Avg. Predicted Prob. \\ 
              \midrule
Male &  10.3\% & 11.5\% \\
Female & 9.8\%  & 13.9\% \\ \bottomrule
\end{tabular}
\caption{\label{tab:gpr-probs} Probability of pronoun referring to neither entity in a sentence of GPR}
\end{table}

\paragraph{Results}  
Although the number of male pronouns associated with no entities in the training data is slightly larger, the model predicted the female pronoun referring to no entities with a significantly higher probability ($p=0.007$ on a permutation test); see \Tref{tab:gpr-probs}. As the training set is balanced, we attribute this bias to the underlying BERT representations. 

We also investigate the relation between the topic of the sentence and model's ability to associate the female pronoun with no entity. We first extracted 20 major topics from the dataset using non-negative matrix factorization \cite{nmf} (refer to Appendix for the list of topics). We then compute the bias score for each topic as the sum of the \emph{log probability bias} score for the top 15 most prevalent words of each topic weighted by their weights within the topic. For this, we use a generic template ``[TARGET] are interested in [ATTRIBUTE]" where TARGET is either men or women. Next we compute a bias score for each sample in the training data as the sum of individual bias scores of topics present in the sample, weighted by the topic weights. Finally, we measured the Spearman correlation coefficient to be 0.207 (which is statistically significant with $p=4e-11$) between the bias scores for male gender across all samples and the model's probability to associate a female pronoun with no entity. We conclude that models using BERT find it challenging to perform coreference resolution when the gender pronoun is female and if the topic is biased towards the male gender.       


\section{Real World Implications}
\label{sec:implications}
In previous sections, we discussed that BERT has human-like biases, which are propagated to downstream tasks. In this section, we discuss another potential negative impact of using BERT in a downstream model. Given that three quarters of US employers now use social media for recruiting job candidates \cite{segal2014social}, many applications are filtered using job recommendation systems and other AI-powered services. \citet{zhao2018learning} discussed that resume filtering systems are biased when the model has strong association between gender and certain professions. Similarly, certain gender-stereotyped attributes have been strongly associated with occupational salary and prestige \cite{glick1991trait}. Using our proposed method, we investigate the gender bias in BERT embeddingss for certain occupation and skill attributes.


\noindent\textbf{Datasets}: 
We use three datasets for our study of gender bias in employment attributes: 

\begin{itemize}
    \item \textit{Employee Salary Dataset\footnote{\url{https://catalog.data.gov/dataset/employee-salaries-2017}} for Montgomery County of Maryland-} Contains 6882 instances of ``Job Title" and ``Salary" records along with other attributes. We sort this dataset in decreasing order of salary and take the first 1000 instances as a proxy for high-paying and prestigious jobs. 
    \item \textit{Positive and Negative Traits Dataset\footnote{\url{http://ideonomy.mit.edu/essays/traits.html}}-} Contains a collection of 234 and 292 adjectives considered ``positive" and ``negative" traits, respectively.
    \item \textit{O*NET 23.2 technology skills\footnote{\url{https://www.onetcenter.org/database.html\#individual-files}}} Contains 17649 unique skills for 27660 jobs, which are posted online
\end{itemize}

\noindent\textbf{Discussion} 
We used the following two templates to measure gender bias: 
\begin{itemize}
    \item ``TARGET is ATTRIBUTE", where TARGET are male and female pronouns viz. \textit{he} and \textit{she}. The ATTRIBUTE are job titles from the Employee Salary dataset, or the adjectives from the Positive and Negative traits dataset.
    \item ``TARGET can do ATTRIBUTE", where the TARGETs are the same, but the ATTRIBUTE are skills from the O*NET dataset.
\end{itemize}

Table \ref{tab:employment} shows the percentage of attributes that were more strongly associated with the male than the female gender. The results prove that BERT expresses strong preferences for male pronouns, raising concerns with using BERT in downstream tasks like resume filtering. 


\begin{table}[]
\small
\centering
\begin{tabular}{lr}
\toprule
Dataset    & Percentage \\ \midrule
Salary     & 88.5\%     \\
Pos-Traits & 80.0\%       \\
Neg-Traits & 78.9\%     \\ 
Skills     & 84.0\%       \\\bottomrule 
\end{tabular}
\caption{\label{tab:employment} Percentage of attributes associated more strongly with the male gender}
\end{table}

\section{Related Work} 
\label{sec:related_work}

NLP applications ranging from core tasks such as coreference resolution \citep{rudinger2018gender} and language identification \cite{jurgens2017incorporating}, to downstream systems such as automated essay scoring \cite{amorim2018automated}, 
exhibit inherent social biases which are attributed to the datasets used to train the embeddings \cite{barocas2016big,zhao2017men,yao2017beyond}. 
There have been several efforts to investigate the amount of intrinsic bias within uncontextualized word embeddings in binary \cite{bolukbasi2016man,garg2018word,swinger19bias} and multiclass \cite{manzini19multiclass} settings. 



Contextualized embeddings such as BERT \cite{devlin2018bert} and ELMo \cite {peters2018deep} have been replacing the traditional type-level embeddings. 
It is thus important to understand the effects of biases learned by these embedding models on downstream tasks. However, it is not straightforward to use the existing bias-exposure methods for contextualized embeddings. For instance, \citet{may2019measuring} used WEAT on sentence embeddings of ELMo and BERT, but there was no clear indication of bias. Rather, they observed counterintuitive behavior like vastly different $p$-values for results concerning gender. 

Along similar lines, \citet{basta2019evaluating} noted that contextual word-embeddings are less biased than traditional word-embeddings. Yet, biases like gender are propagated heavily in downstream tasks. For instance, \citet{ZWYCOC19} showed that ELMo exhibits gender bias for certain professions. As a result, female entities are predicted less accurately than male entities for certain occupation words, in the coreference resolution task. \citet{field19personabias} revealed biases in ELMo embeddings that limit their applicability across data domains. 
Motivated by these recent findings, our work proposes a new method to expose and measure bias in contextualized word embeddings, specifically BERT. As opposed to previous work, our measure of bias is more consistent with human biases. We also study the effect of this intrinsic bias on downstream tasks, and highlight the negative impacts of gender-bias in real world applications.

\section{Conclusion} 

In this paper, we showed that querying the underlying language model can effectively measure bias in BERT and expose multiple stereotypes embedded in the model. We also showed that our measure of bias is more consistent with human-biases, and outperforms the traditional WEAT method on BERT. Finally we showed that these biases can have negative downstream effects. In the future, we would like to explore the effects on other downstream tasks such as text classification, and device an effective method of debiasing contextualized word embeddings.


\section*{Acknowledgments}
This material is based upon work supported by the National Science Foundation under Grant No.~IIS1812327.

\bibliography{acl2019}
\bibliographystyle{acl_natbib}

\newpage

\section*{Appendix}
\begin{table} [hbt]
\small
\begin{center}
\begin{tabular}{lp{18em}}
\hline 
Topic Id & Top 5 Words \\
\hline
1 & match,round,second,team,season \\
2 & times,city,jersey,york,new \\
3 & married,son,died,wife,daughter \\
4 & best,award,actress,films,film \\
5 & friend,like,work,mother,life \\
6 & university,music,attended,high,school \\
7 & president,general,governor,party,state \\
8 & songs,solo,song,band,album \\
9 & medal,gold,final,won,world \\
10 & best,role,character,television,series \\
11 & kruse,moved,amy,esme,time \\
12 & usa,trunchbull,pageant,2011,miss \\
13 & american,august,brother,actress,born \\
14 & sir,died,church,song,john \\
15 & natasha,days,hospital,helene,later \\
16 & played,debut,sang,role,opera \\
17 & january,december,october,july,married \\
18 & academy,member,american,university,family \\
19 & award,best,played,mary,year \\
20 & jersey,death,james,king,paul \\

\hline
\end{tabular}
\end{center}
\caption{\label{gpr-topics} Extracted topics for the GPR dataset}
\end{table}

\end{document}